# Neural Network Clustering Based on Distances Between Objects

Leonid B. Litinskii, Dmitry E. Romanov

Institute of Optical-Neural Technologies Russian Academy of Sciences, Moscow
litin@iont.ru, demaroman@yandex.ru

**Abstract.** We present an algorithm of clustering of many-dimensional objects, where only the distances between objects are used. Centers of classes are found with the aid of neuron-like procedure with lateral inhibition. The result of clustering does not depend on starting conditions. Our algorithm makes it possible to give an idea about classes that really exist in the empirical data. The results of computer simulations are presented.

## 1. Introduction

Data clustering deals with the problem of classifying a set of $N$ objects into groups so that objects within the same group are more similar than objects belonging to different groups. Each object is identified by a number $m$ of measurable features, consequently, $i$th object can be represented as a point $\mathbf{x}_i \in \mathbf{R}^\mathbf{m}$, $i = 1, 2, ..., N$. Data clustering aims at identifying clusters as more densely populated regions in the space $\mathbf{R}^\mathbf{m}$.

This is a traditional problem of unsupervised pattern recognition. During last 40 years a lot of approaches to solve this problem were suggested. The general strategy is as follows: at first, somehow or other one finds the optimal partition of the points into $K$ classes, and then changes the value of the parameter $K$ from $N$ to 1. Here the main interest is the way how small classes (relating to big values of $K$) are combined into bigger classes (relating to small values of $K$). These transformations allow us to get some idea about the structure of empirical data. They indicate mutual location of compact groups of points in many-dimensional space. They also indicate which of these groups are close and which are far from each other. Interpretation of the obtained classes in substantial terms, and it is no less important, the details of their mutual location allows the researcher to construct meaningful models of the phenomenon under consideration.

Different methods of data clustering differ from each other by the way of finding of the optimal partition of the points into $K$ classes. It is literally to say that almost all of them own the same poor feature: the result of partition into $K$ classes depends on arbitrary chosen initial conditions, which have to be specify to start the partition procedure. Consequently, to obtain the optimal partition, it is necessary to repeat the procedure many times, each time starting from new initial conditions. In general, it cannot be guaranteed that the optimal partition into $K$ classes would be found. Here the situation is close to the one, which we face when founding the global minimum of multiextremal functional. The problems of such a kind exhibit a tendency to become NP-complete. This means that for large $N$ only a local optimal partition can be found, but not necessary the best one.

Thus, almost all clustering methods based on the local partition of objects into $K$ classes. Among them there are the well-known and most simple *K-means approach* [1]-[3], mathematically advanced *Super-Paramagnetic Clustering* [4] and *Maximum Likelihood Clustering* [5], popular in Russia *the FOREL-type* algorithms [3] and prevailing in the West different variants of *Hierarchical Clustering* [6]. Let us show the problem, using two last approaches as examples.

The general scheme of the *FOREL*-algorithm is as follows: 1) we specify a value $T$ that is the radius of $m$-dimensional sphere, which in what follows is used as a threshold for interaction radius between points; 2) we place the center of the sphere with the radius $T$ at an arbitrary input point; 3) we find coordinates of the center of gravity of points that find themselves inside the sphere; 4) we transfer the center of the sphere in the center of gravity and go back to item 3; 5) as far as when going from one to the next iterating the sphere remains in the same place, we suppose that the points inside it constitute a class; we move them away from the set and go back to the item 2.

It is clear that after finite number of steps we obtain a partition of the points into some classes. In each class the distances between points are less than $2T$. However, the result of partition depends on the starting point, where the center of the sphere is situated (see item 2). Since the step 2 is repeated again and again, it is evident that the number of different partitions (for fixed $T$) can be sufficiently large.

The Hierarchical Clustering is based on a very simple idea too. Given some partition into $K$ classes, it merges the two closest classes into a single one. So, starting from the partition into $K = N$ classes, the algorithm generates a sequence of partitions as $K$ varies from $N$ to 1. The sequence of partitions and their hierarchy can be represented by a dendrogram. Applications of this approach are discussed for example in [6]. However, there is no explanation why just two closest classes have to be combined. After all, this is only one of possible reasonable recipes. This leads to local optimal partition too.

Of course, there is no reason to dramatize the situation with regard to local optimality of partitions. During 40 years of practice, approaches and methods were developed allowing one to come to correct conclusions basing on local optimal partitions of objects into classes. However, it is very attractive to construct a method, which would not depend on an arbitrary choice of initial conditions. Just such an algorithm is presented in this publication.

The same as the *FOREL*-algorithm our method is based on introduction of an effective interaction radius $T$ between points $\mathbf{x}_i$ and the partition of the points between spheres of radius $T$. The points that get into a sphere belong to one class. The centers of the spheres are obtained as a result of a neuron-like procedure with lateral inhibition. As a result, the centers of the spheres are the input points, which for a given radius $T$ interact with maximal number of surrounding points. It can be said that the centers of the spheres are located inside the regions of concentration of the input points. At the same time, we determine the number of classes $K$ that are characteristic for the input data for a given interaction radius $T$. Then, the value $T$ changes from zero to a very large value. We plot the graph *K(T)*, which shows the dependence of the number of classes on *T*. We can estimate the number of real classes that are in the empirical data by the number of lengthy "plateau" on this graph. The calculation complexity of the algorithm is estimated as $O(N^2)$.

In the present publication we describe the clustering algorithm and the results of it testing with the aid of model problems and empirical data known as "Fisher's irises" [2].

## 2. Clustering algorithm

For a given set of *m*-dimensional points $\{\mathbf{x}_i\}_1^N \in \mathbf{R}^\mathbf{m}$ we calculate a quadratic $(N \times N)$-matrix of Euclidean distances between them: $\mathbf{D} = (D_{ij})_{i,j=1}^N$. In what follows we need these distances $D_{ij}$ only. We suppose that in each point $\mathbf{x}_i$ there is a neuron with initial activity $S_i(0)$, which will be defined below.

**1)** For a fixed interaction threshold $T > 0$ let us set the value of a connection $w_{ij}$ between *i*th and *j*th neurons as

$$w_{ij} = \begin{cases} \dfrac{T^2}{D_{ij}^2 + T^2}, & \text{when } D_{ij} \leq T, \\ 0, & \text{when } D_{ij} > T. \end{cases}$$

As we see, there are no connections between neurons, if the distance between points is greater than *T*. Note, $w_{ii}(T) \equiv 1 \; \forall \, i$.

**2)** Let us set initial activity of each neuron to be

$$S_i(0) = \sum_{j=1}^N w_{ij} \geq 1$$

Neurons, which are inside agglomerations of the points, have large initial activity, because they have more nonzero connections than neurons at the periphery of agglomerations.

**3)** We start the activities "transmitting" process:

$$S_i(t+1) = S_i(t) + \alpha \sum_{j=1}^N w_{ij}(S_i(t) - S_j(t))$$

where $\alpha$ is the parameter characterizing the transmitting speed. It is easy to see that during the transmitting process a neuron with large initial activity "takes away" activities from neurons with whom it interacts and whose activities are less. The activities of these surrounding neurons decrease steadily.

**4)** If during the transmitting process the activity of a neuron becomes negative $S_i(t) < 0$, we set $S_i \equiv 0$, and eliminate this neuron from the transmitting process (it has nothing to give away).

It is clear that little by little the neurons from the periphery of agglomerations shall drop out giving away their activities to neurons inside the agglomerations. This means that step by step the neurons from the periphery will leave the field. Gradually, we shall have a situation, when only some far from each other non-interacting neurons with nonzero activities remain. Subsequent transmitting is impossible and the procedure stops.

**5)** Suppose as a result of the transmitting process *K* neurons remain far away from each other. The input points $\mathbf{x}_i$ corresponding to these neurons will be called the centers of the classes. All other input points $\mathbf{x}_j$ are distributed between classes basing on the criterion of maximal closeness to one or another center.

The items 1)-5) are carried out for a fixed interaction threshold *T*. It is clear that if $T \approx 0$, no one neuron interacts with another one ($w_{ij} = 0$ when $i \neq j$). All the neurons have the same activities $S_i(0) = w_{ii} = 1$. No transmitting process will have place. So we get a great number *N* of classes, each of which consists of one input point only. On the other hand, if the interaction threshold *T* is very large (for example, it is greater than $\max(D_{ij})/2$) all neurons are interacting with each other, and as the result of transmitting only one neuron remains active. We can say that it is located inside "the cloud" of the input points. In this limiting case there is only one class including all the input points.

Changing *T* from zero to it maximal value, we plot the dependence of the number of classes on the value of the interaction threshold, $K(T)$. It was found that for these graphs the presence of long, extensive "plateaus" is typical. In other words, the number of classes *K* does not change for some intervals, where *T* changes. These "plateaus" allows one to estimate the number of classes existing really in empirical data (for the first time this criterion was proposed by the author of [3]).

## 3. The results of computer simulations

The simplest case is shown in Fig.1. On the left panel we see 50 points distributed on the plane into five clusters, on the right panel the obtained graph *K(T)* is shown. We see that at the initial stage of changing of *T* the number of classes changes very rapidly. Then it becomes stabilize at the level *K* = 5 and does not change in the some interval of changing of *T*. Here empirical points are distributed into 5 input clusters exactly. The next plateau corresponds to *K* = 4. In this case the points that belong to two close internal agglomerations combine into one class. The next plateau is when *K* = 3; here the points belonging to two pairs of close to each other agglomerations are combined into two classes. Finally, the last plateau is at the level *K* = 2; here all the points from the lower part of the figure are combined into one class.

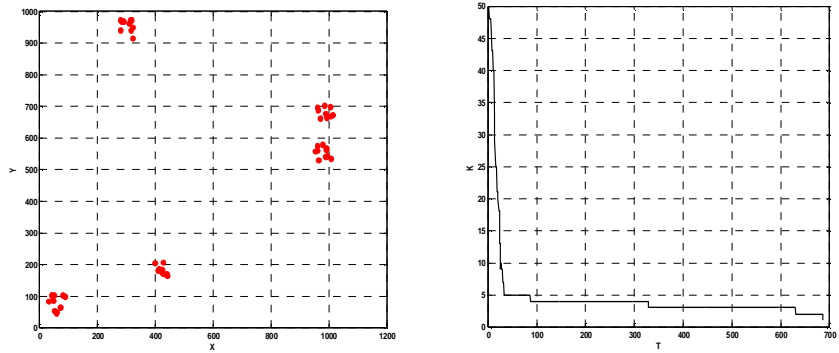

**Fig. 1.** On the left panel there are 5 compact clusters formed by 50 points at the plane (X,Y); on the right panel the graph $K(T)$ is presented.

In the case of 10 classes (see Fig.2) on the graph small plateaus can be observed when $K$ = 10, 9, 8, 7, 5, 4. However, the widest plateau we observe when $K$ = 2. This plateau corresponds to the partition of the points between two classes that are sufficiently clearly seen in the left panel of the figure.

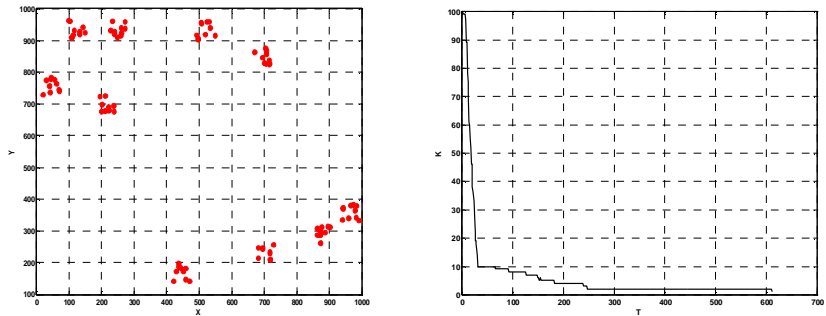

**Fig. 2.** The same as in the previous figure for 100 points and 10 compact clusters.

The last example is the classical «Fisher's irises» clustering problem. The input data is a set of 150 four-dimensional points (four parameters characterize each iris). It is known that these 150 points are distributed between 3 compact classes. Two of them are slightly closer to each other than the third. Our algorithm showed just the same picture of irises distribution between classes (Fig.3). The first plateau is at the level $K$=3, and the next plateau corresponds to $K$=2.

These examples demonstrate that of interest is not only the separation of the objects into a certain number of classes, but also the way in which small classes join into larger ones. These transformations indicate which of small compact classes of objects are close to each other and allow one to understand the intrinsic structure of empirical data.

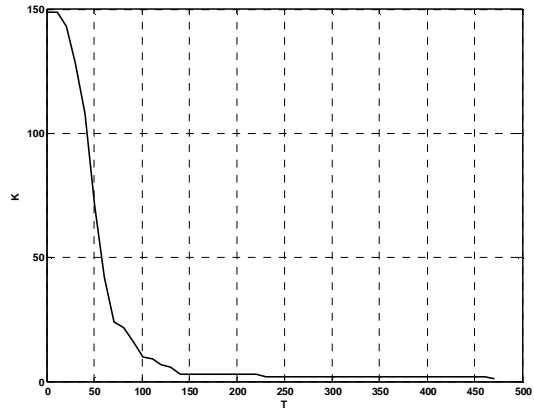

**Fig. 3.** The graph for Fisher's irises.

## 4. Discussion

In real empirical data in addition to compact agglomerations certainly there are "noisy" points, which can be found in the space between the agglomerations. In this case our algorithm gives plateaus when the threshold $T$ is small. The number of classes $K$ corresponding to these values of $T$ is sufficiently large: $K\gg 1$. These classes are mostly fictitious ones. Each of them consists from one noisy point only.

   The reason is clear: when the value of the threshold $T$ is small each noisy point has a chance to make its own class consisting from this one point only. For beginning values of $T$ some noisy points take their occasion. That is why there are short, but distinct plateaus corresponding to the large number of classes $K$. It is easy to ignore these fictitious classes: it is enough to take into account only classes where there are a lot of points. Thus, our algorithm is able to process noisy data sets.

   The algorithm is based on the hypothesis of a compact grouping points in the classes and is oriented on the unsupervised pattern recognition. In a general case it is unable to process overlapping sets of data belonging to different classes. This is a problem of supervised pattern recognition. It is possible, however, that different classes overlap slightly, only by their periphery. Then our algorithm can be modified, so that it can separate cores of the classes. Such an approach can be of considerable use, and we plan to examine it in details.

Our algorithm has a usefull property: it works not only for points $\{\mathbf{x}_i\}_1^N$ in the coordinate presentation, but also when we know the distances between points only. These two types of input data, the coordinate presentation of the points, on the one hand, and the distances between the points, on the other, are not equivalent. Indeed, if only the distances between the points are known, it is not so simple to reconstruct the coordinates of the points themselves. As a rule, this can be done only under some additional assumptions, and these assumptions have an influence on the solution of the problem.

Still more general is the clustering problem, when the input data is an arbitrary symmetrical matrix (not necessary the matrix of distances between the points). In this case matrix elements can be negative. As a rule the diagonal elements of such a matrix are equal to zero. Then ($ij$)th matrix element is treated as a measure of connection between $j$th and $i$th objects. The clustering of this matrix aims at finding of strongly connected groups of objects. Of course, matrix elements can be normalized in such a way that they can be treated as "scalar products" of vectors. However, the problem is that with the aid of these "scalar products" it is impossible to reconstruct uniquely even the distances between the vectors. Consequently, to solve the most general clustering problems our algorithm has to be modified. We plan to do this in the following publication.

The work was supported by Russian Basic Research Foundation (grants 05-07-90049 and 06-01-00109).